\documentclass[conference]{IEEEtran}
\IEEEoverridecommandlockouts
\usepackage{cite}

\usepackage{amsmath,amssymb,amsfonts}
\usepackage{algorithmic}
\usepackage{graphicx}
\usepackage{textcomp}
\usepackage{xcolor}
\def\BibTeX{{\rm B\kern-.05em{\sc i\kern-.025em b}\kern-.08em
    T\kern-.1667em\lower.7ex\hbox{E}\kern-.125emX}}

\usepackage{subfig}
\usepackage{enumitem}
\usepackage{hyperref}
\usepackage{relsize}
\usepackage{tikz}
\usepackage{pgfplots}
\pgfplotsset{compat=newest}
\usepackage{float}
\usepgfplotslibrary{colorbrewer}

\begin{document}

\title{A Game Interface to Study Semantic Grounding in Text-Based Models
\thanks{This work was supported  by a public grant overseen by the French National Research Agency (ANR) as part of the ``Investissements d'Avenir'' program: Idex \emph{Lorraine Universit\'e d'Excellence} (reference: ANR-15-IDEX-0004).}
}

\author{\IEEEauthorblockN{Timothee Mickus}
\IEEEauthorblockA{\textit{ATILF} \\
\textit{CNRS, Université de Lorraine}\\
Nancy, France \\
\tt tmickus@atilf.fr}
\and
\IEEEauthorblockN{Mathieu Constant}
\IEEEauthorblockA{\textit{ATILF} \\
\textit{CNRS, Université de Lorraine}\\
Nancy, France \\
\tt mconstant@atilf.fr}
\and
\IEEEauthorblockN{Denis Paperno}
\IEEEauthorblockA{\textit{Linguistics Department} \\
\textit{Utrecht University}\\
Utrecht, Netherlands \\
\tt d.paperno@uu.nl}
}

\IEEEpubid{
\begin{minipage}{\textwidth}\ \\[12pt]
978-1-6654-3886-5/21/\$31.00 \copyright 2021 European Union
\end{minipage}}

\maketitle

\begin{abstract}
Can language models learn grounded representations from text distribution alone?
This question is both central and recurrent in natural language processing; authors generally agree that grounding requires more than textual distribution.
We propose to experimentally test this claim: if any two words have different meanings and yet cannot be distinguished from distribution alone, then grounding is out of the reach of text-based models.
To that end, we present early work on an online game for the collection of human judgments on the distributional similarity of word pairs in five languages.
We further report early results of our data collection campaign.
\end{abstract}


\section{Introduction} \label{sec:intro}

Semantic grounding is a central question in Natural Language Processing.
Studies on grounding try to see how words and sentences can be linked to real-world objects.
This question has recently gained prominence through Bender \& Koller \cite{octopus}, who specifically set out to disprove that large language models like BERT \cite{bert} or GPT-3 \cite{gpt-3} are capable of building semantically coherent representations.
The thought experiment of Bender \& Koller echoes previous works such as Harnad's Symbol Grounding Problem \cite{chinese-dict} or Searle's Chinese Room Argument \cite{chinese-room-argument}. 

The prominence of thought experiments, rather than experimental procedures, shows how elusive this question is.
That is not to say that no experimental approach has been adopted: multimodal evaluation benchmarks exist \cite{vqa}, and authors have proposed semantically grounded models of language \cite{vis,audio,olfac}.
Yet most researchers seemingly adopt without discussion the position that something more than text is required to coherently describe the world.
Here, we propose an experiment to test this claim.
If there are pairs of words that humans cannot distinguish from distribution alone, then in principle text-based models cannot consistently and systematically use words in semantically appropriate contexts---because even perfectly grounded semantic representations as we expect to find in humans do not suffice to solve the task.
This can also be seen as an investigation of the limits of the distributional hypothesis \cite{harris}: we are looking into documenting cases where distribution is not enough to infer meaning.

We first study how to formulate a task that will be natural both for humans and distributional semantics models in Section~\ref{sec:cstp}.
We then detail our game-based data collection campaign in Section~\ref{sec:game}.
We present early results in the form of usage statistics in Section~\ref{sec:stats}.
Lastly, we give some concluding remarks in Section~\ref{sec:ccl}.

\section{Distributional Models of Word Meaning} \label{sec:cstp}

Most modern language models and embedding architectures used in NLP can be linked to Zellig Harris' distributional hypothesis \cite{harris}.
These models, while they all differ in architecture, share the common feature that they can exhibit which of two words they prefer in a specific context, or formally:
\begin{equation} \label{eq:cstp}
    \Pr(t_1|c) > \Pr(t_2|c)     
\end{equation}

Some distributional models directly model $P(t | c)$. In particular, we can quote sequence denoising objectives \cite{bert,lewis-etal-2020-bart} or multinomial classification objectives like \textsc{cbow} \cite{mikolov-2013-efficient}.
Other architectures like skip-gram \cite{mikolov-2013-efficient} model $P(c|\cdot)$.
In such cases, finding some equation corresponding to eq.~\eqref{eq:cstp} can be done by a straightforward application of Bayes' rule:
\begin{align}
    P(t_1 |c) > P(t_2 | c)  &= \frac{P(c|t_1)P(t_1)}{P(c)} > \frac{P(c|t_2)P(t_2)}{P(c)} \nonumber \\
                            &= P(c|t_1)P(t_1) > P(c|t_2)P(t_2)
\end{align}
A third family of architectures that can be linked to this framework concern negative sampling approaches \cite{bojanowski-2016-enriching,clark-etal-2020-pre}, which are trained to estimate $P(t \in c)$.
To arrive exactly at the previous eq.~\eqref{eq:cstp}, we can re-normalize this probability distribution with respect to the entire vocabulary:
\begin{equation*}
	\Pr(t_i | c) = \frac{1}{\sum \limits_{t_j \in V} P(t_j \in c)} \cdot P(t_i \in c)
\end{equation*}
The above does sum to 1 over the full probability space of the vocabulary $V$, is defined with respect to the context $c$. 
As $\frac{1}{\sum \limits_{t_j \in V} P(t_j \in c)}$ is constant for a given context $c$, negative sampling models therefore yield the following formulation:
\begin{equation}
    \Pr(t_1|c) > \Pr(t_2|c) = P(t_1 \in c) > P(t_2 \in c)
\end{equation}
Lastly, generative models like \cite{bengio-etal-2003-neural,gpt-3} yield the probability of $t_{i}$ being the next token: $P(t_{i} | c_{1 \dots i-1})$. 
Hence, works like \cite{linzen-2016} derive the comparison given the full context $c_{1 \dots n}$ as:
\begin{align}
    &P(c_n | c_{1 \dots i-1}, ~ t_1, ~ c_{i+1 \dots n-1}) \nonumber \\
    &\qquad \qquad > P(c_n | c_{1 \dots i-1}, ~ t_2, ~ c_{i+1 \dots n-1})
\end{align}

This Context-Specific Term Preference (CSTP) therefore allows us to evaluate all these models using a united framework, specifically by answering which of two words is preferable in a given context.
This entails that we can compare various architectures on this task.
Crucially, the task is also easily understood by humans, and we therefore can collect humans judgments to compare distributional models to.

\section{Game Interface} \label{sec:game}

\begin{figure}[!t]
    \centering
    \includegraphics[width=\linewidth]{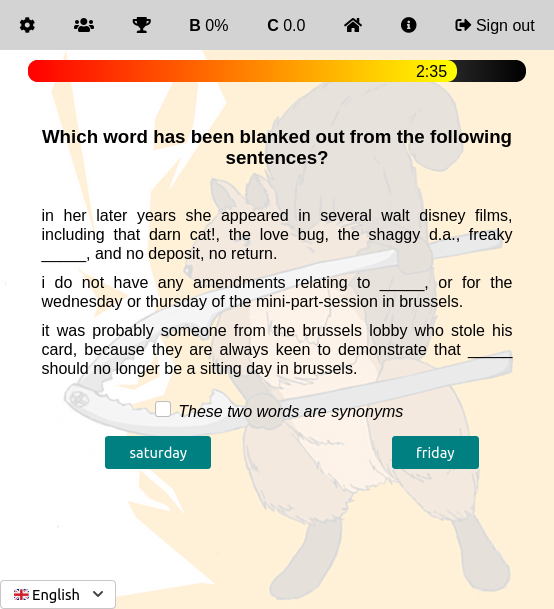}
    \caption{Game Annotation Interface}
    \label{fig:interface}
\end{figure}

We propose to collect such human judgments through a game interface, as displayed in Figure~\ref{fig:interface}. 
This game is available online at \url{https://blankcrack.atilf.fr/}.\footnote{
    Code for the interface will be made public at a later date.
} 

There are two aspects in which human knowledge can be useful to our enterprise:
\begin{enumerate}[label=(\alph*)]
    \item \label{q:cracker} How hard is it to distinguish terms from contexts alone?
    \item \label{q:blanker} Are there terms indistinguishable from context alone?
\end{enumerate}
These two related questions lead to two distinct series of data to collect.
To answer \ref{q:cracker}, we would need to run the CSTP task, but substituting distributional models with human annotators.
To answer \ref{q:blanker}, we would instead require human participants to suggest word pairs that they expect to be difficult to distinguish.
These two collection procedures lead us to an adversarial data collection project, where annotators can play either of two roles: proposing word pairs to answer \ref{q:blanker} (henceforth \ref{q:blanker}-annotators), or distinguishing word pairs proposed in \ref{q:blanker} to answer \ref{q:cracker} (likewise \ref{q:cracker}-annotators).

Once the data are collected, we will be able to compare the performance of language models and distributional models on the very same task (as outlined in Section~\ref{sec:cstp}). 
Pairs that humans find challenging should be difficult for neural models as well---as long as contexts are not attested in these models' training data. 
We expect these collected data to constitute a strong human benchmark for distributional models.

\subsection{Sentence Corpora}

As we are interested in establishing a widely applicable benchmark, we collect data for multiple languages: English, French, Italian, Spanish and Russian.\footnote{
    These five languages were chosen on criteria of high data availability. We intend to facilitate adding new languages to the interface in future releases.
}
The first element required for our game is a list of distributional contexts, or sentences. 
We further wish our data to be broadly comparable across languages: we therefore select a comparable number of sentences from comparable but varied corpora.
We chose to select 4M sentences per language, equally drawn from four genres of corpora: Wikipedia dumps, books corpora (Gutenberg Project, WikiSource, \url{LiberLiber.it}), parliamentary debates (EuroParl \cite{europarl} \&  UN Corpus \cite{un-corpus}) and subtitles (OpenSubtitles \cite{opus}). 

\subsection{Initial Word Pairs Set} \label{sec:game:init-wp}

The second type of data we require is a set of word pairs to bootstrap our data collection process, so that both \ref{q:cracker}-annotators and \ref{q:blanker}-annotators can immediately start.
To do so, we consider two strategies.
The first ``manual'' strategy consists in manually constructing pairs that one initially expects to be challenging, such as months, days of the week, numbers (cardinal and ordinal) and colors. 
Any pair of terms from one of these series can constitute a word pair to annotate.
The second strategy, which we call ``\textsc{w2v}-based'', consists in automatically discovering distributionally similar items given our corpus of sentence.
We train distinct word2vec models for each of our five language-specific corpora.\footnote{
    We select hyperparameters with Bayesian optimization \cite{bayesopt}, using performance on a formal analogy dataset as the objective to maximize.
}
We randomly sample 1M word pairs, and narrow down to the 250 items whose vectors maximize cosine similarity.\footnote{
    We considered alternative strategies, such as using ontologies like WordNet to select initial word pairs, but distributional similarity proved the most efficient during a short pilot study.
}

\subsection{Annotation Items}

From these word pairs and sentences, we can then automatically construct annotation items. 
We present each \ref{q:cracker}-annotator with two words $w_a$ and $w_b$ from a given word pair $\langle w_a, w_b \rangle$, and $k$ sentences\footnote{
     \ref{q:cracker}-annotators can set $k \in \{1, 3, 5\}$; by default, $k=5$.
} randomly selected such that all sentences contain the {target} $w_a$ and none contains a word with the same word stem as the {foil} $w_b$.
We replace all occurrences of $w_a$ by a blank token before presenting them to the \ref{q:cracker}-annotator. 
The annotator is then tasked with retrieving which of the target $w_a$ or the foil $w_b$ corresponds to these blank tokens.
Word pairs $\langle w_a, w_b \rangle$ can correspond either to our initial set of word pairs, or to items proposed by \ref{q:blanker}-annotators.

\subsection{Player Engagement} \label{sec:game:engagement}

At its core, our game is score-based, with two distinct scores per user corresponding to performances as \ref{q:cracker}-annotators and \ref{q:blanker}-annotators. 
The \ref{q:blanker}-annotator score corresponds to the success rate (as a percentage) of the user's proposed word pairs, i.e., how often \ref{q:cracker}-annotators failed to solve riddles constructed using the \ref{q:blanker}-annotator's word pairs, and selected the foil $w_b$ instead of the target $w_a$.
The \ref{q:cracker}-annotator score is a running tally of points: \ref{q:cracker}-annotator get between 0.1 and 3 points per correctly solved annotation item (where the \ref{q:cracker}-annotator selected the target $w_a$, rather than the foil $w_b$), depending on whether the item was solved under 3 minutes, was based on a known difficult pair, or whether the \ref{q:cracker}-annotator had set a lower number $k$ of example sentences.

The possibility to set the number $k$ of sentences per riddle is presented in-game as a difficulty level setting.
Aside from this setting, we further implement several mechanisms to attempt to keep players engaged.
First, we include a competition mode, whereby users compete against one another; this competition mode is based on a ``friends list''.
Second, we ensure that word pairs newly suggested by \ref{q:blanker}-annotators get presented to \ref{q:cracker}-annotators in priority, so that \ref{q:blanker}-annotators receive feedback as early as possible.
Third, we also include some materials to share on social media, e.g., when \ref{q:cracker}-annotators successfully retrieve the blanked-out word in their annotations multiple times in a row, or at the end of a competition session.
Thus far, sharing on social media and competitions have not been used much often by our users.

We also note that users tend to connect only once. 
One explanation may lie in that ``manual'' word pairs from our initial set (cf. Subsection~\ref{sec:game:init-wp}) are felt to be very hard to solve. 
We are currently investigating mechanisms to combat this trend such as high-score leader-boards displaying username, language and score for top players; our intuition is that it may motivate players to return to the platform to ensure they still appear on the leader-board. 
Another possibility would be to provide users with a way to opt-out of these word pairs.

\section{Usage statistics} \label{sec:stats}

Thus far, we have collected slightly over 7000 annotations in the first month of widespread advertisement.
We take this as an encouraging response, although improvements in the interface and game engagement mechanisms can be made.

\begin{figure}[t!]
    \centering
    \subfloat[\label{fig:annots:counts} Counts of \ref{q:cracker}-annotations by language.]{
        \begin{tikzpicture}
            \tikzstyle{every node}=[font=\tiny]
            \begin{axis}[
                ybar, ymin=0, ymax=4000,
                symbolic x coords={EN,ES,FR,IT,RU},
                enlarge x limits=0.125,bar width=.22cm,
                enlarge y limits={upper=0.5},
                xtick=data,
                xticklabels={\larger\tt en, \larger\tt es, \larger\tt fr, \larger\tt it, \larger\tt ru},
                nodes near coords, 
                nodes near coords align={anchor=south},
                label style={font=\tiny},
                nodes near coords always on top/.style={
                    every node near coord/.append style={
                        anchor=west,
                        rotate=90,
                        font=\tiny,
                        xshift=-0.5ex,
                        color=black,
                    },
                },
                nodes near coords always on top,
                /pgf/number format/use comma,
                /pgf/number format/1000 sep={},
                height=5.5cm,
                width=\linewidth,
                xtick align=inside,
                cycle list/RdYlBu-5,
                every axis plot/.append style={
                    fill,
                },
            ]
                \addplot+ [draw=black,] coordinates {
                    ({EN},2098)
                    ({ES},1415)
                    ({FR},3629)
                    ({IT},101)
                    ({RU},19)
                };
                \addplot+ [draw=black,] coordinates {
                    ({EN},726)
                    ({ES},556)
                    ({FR},930)
                };
                \addplot+ [draw=black,] coordinates {
                    ({EN},694)
                    ({ES},650)
                    ({FR},1967)
                    ({IT},68)
                    ({RU},15)
                };
                \addplot+ [draw=black,] coordinates {
                    ({EN},678)
                    ({ES},209)
                    ({FR},732)
                    ({IT},33)
                    ({RU},4)
                };
                \legend{all pairs,\ref{q:blanker}-annots,manual,\textsc{w2v}-based}
            \end{axis}
        \end{tikzpicture}
    }
    
    \subfloat[%
            \label{fig:annots:success}%
            Overall success rate of \ref{q:cracker}-annotators (in \%)%
            \protect\footnotemark%
        ]{%
        \begin{tikzpicture}
            \tikzstyle{every node}=[font=\tiny]
            \begin{axis}[
                ybar, ymin=0, ymax=100,
                symbolic x coords={EN,ES,FR,IT,RU},
                bar width=.22cm,
                xtick=data,
                xticklabels={\larger\tt en, \larger\tt es, \larger\tt fr, \larger\tt it, \larger\tt ru},
                enlarge x limits=0.125,
                enlarge y limits={upper=0.5},
                legend style={anchor=south east},
                /pgf/number format/use comma,
                /pgf/number format/1000 sep={},
                nodes near coords, 
                nodes near coords always on top/.style={
                    every node near coord/.append style={
                        font=\tiny,
                        color=black,
                    },
                },
                nodes near coords always on top,
                height=5.5cm,
                width=\linewidth,
                legend pos=south east,
                xtick align=inside,
                label style={font=\tiny},
                cycle list/RdYlBu-5,
                every axis plot/.append style={
                    fill,
                },
            ]
                \addplot+ [draw=black,] coordinates {
                    ({EN},79)
                    ({ES},80)
                    ({FR},81)
                    ({IT},79)
                    ({RU},95)
                };
                \addplot+ [draw=black,] coordinates {
                    ({EN},86)
                    ({ES},91)
                    ({FR},86)
                };
                \addplot+ [draw=black,] coordinates {
                    ({EN},77)
                    ({ES},79)
                    ({FR},76)
                    ({IT},76)
                    ({RU},93)
                };
                \addplot+ [draw=black,] coordinates {
                    ({EN},73)
                    ({ES},52)
                    ({FR},88)
                    ({IT},85)
                    ({RU},100)
                };
            \legend{all pairs,\ref{q:blanker}-annots,manual,\textsc{w2v}-based}

            \end{axis}
        \end{tikzpicture}%
    }%
    \caption{Overview of collected \ref{q:cracker}-annotations}
    \label{fig:annots}
\end{figure}
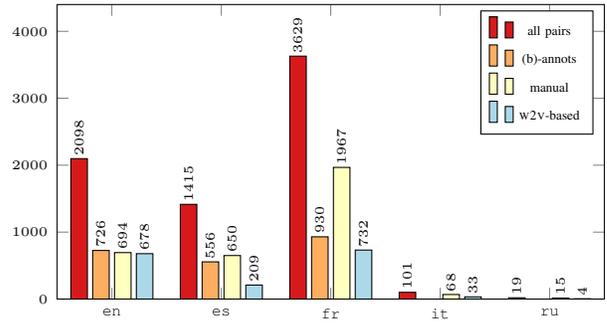
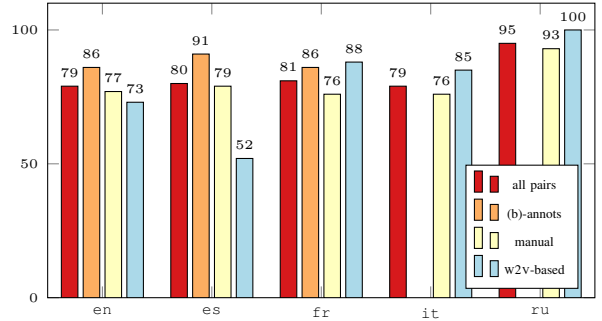

A more precise breakdown is displayed in Figure~\ref{fig:annots}.
Subfigure~\ref{fig:annots:counts} displays the number of annotations we collected, whereas Subfigure~\ref{fig:annots:success} provides an estimate of the success rate of \ref{q:cracker}-annotators. 
Each subfigure is broken down further according to the type of word pairs: ``manually constructed'' and ``\textsc{w2v}-based'' both come from our initial set (see Subsection~\ref{sec:game:init-wp}) whereas ``\ref{q:blanker}-annots'' have been proposed by \ref{q:blanker}-annotators.
The first obvious trend we can observe from Subfigure~\ref{fig:annots:counts} is that the number of Italian and Russian annotations is severely lacking, with respect to English, French and Russian.
For these two languages, word pairs from \ref{q:blanker}-annotators have yet to be annotated.
This low number of annotations indicates that percentages from Subfigure~\ref{fig:annots:success} are most likely off.
This is a consequence of our current advertisement: we have not yet reached out to Italian and Russian audiences.

On average, 80\% of all items are correctly annotated by \ref{q:cracker}-annotators.
\footnotetext{
    No data collected for \ref{q:blanker}-annotators word pairs in Italian and Russian. 
    Russian and Italian percentages are unreliable, owing to the too few items collected.
} 
This suggests the task is in itself rather simple, which would entail that in principle, most words can be distinguished from distribution alone.
On the other hand, this also entails that humans do not invariably solve the task with perfect accuracy, and we intend to strongly focus on incorrect \ref{q:cracker}-annotations in future research.
Likewise, note that a random baseline would already entail a rate of 50\%.

If we compare \ref{q:cracker}-annotator success rates on word pairs from \ref{q:blanker}-annotators and on our two strategies for constructing of word pairs (Subfigure~\ref{fig:annots:success}), we see that the latter are generally found more difficult than the former, corroborating users' feedback (see Subsection~\ref{sec:game:engagement}). 
Another interesting trend to note is that Spanish, French and English annotators seem to differ from one another: word pairs from Spanish \ref{q:blanker}-annotators are found to be generally easier than what we observe in other languages, whereas ``manual'' word pairs are found to be harder across the board, but with varying degrees: Spanish pairs are found to be easier than English pairs, which in turn are easier then French pairs, while our ``\textsc{w2v}-based'' strategy yields the opposite ordering. 
At this stage of our experiments, we cannot rule out an effect of sample selection, especially given the surprisingly low, barely above chance \ref{q:cracker}-annotator success rate for Spanish `\textsc{w2v}-based'' word pairs. 

\begin{figure}[!t]
    \centering
    \includegraphics[width=\linewidth]{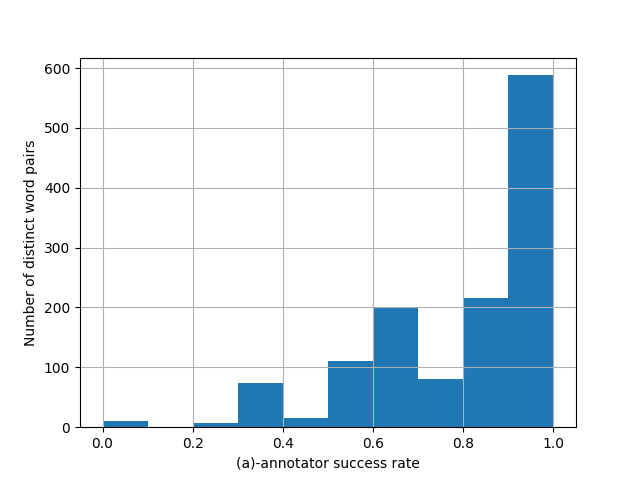}
    \caption{Number of distinct word pairs per \ref{q:cracker}-annotators success rate, all languages combined.}
    \label{fig:hist-data}
\end{figure}

Lastly, Figure~\ref{fig:hist-data} summarizes the number of distinct word pairs for various bins of \ref{q:cracker}-annotators success rate. 
To mitigate noise, we ignore word pairs that have been annotated less than three times: for such pairs, the associated average success rate can only be either 0, 1 or 0.5 if it has been seen twice, which would inflate values at these specific peaks.
We can see that the vast majority of our 1656 distinct word pairs is found to be rather simple to distinguish: 588 pairs are solved correctly 9 times out of ten or more, and 804 are solved correctly 8 times out of ten or more.
This leaves us with a significant number of word pairs with low scores, some around chance level, which we plan to focus on in future experiments.
It is important to note that this illustration is not entirely reliable: the average number of \ref{q:cracker}-annotations per word pair is 4.6, meaning that \ref{q:cracker}-annotators success rates are often not statistically reliable. 
In any event, more data collection and curating is necessary.

\section{Conclusions} \label{sec:ccl}

We have presented an online game platform for collecting human judgments on the distributional hypothesis. 
We have outlined theoretical motivations based on a formal approach subsuming many different distributional models (Section~\ref{sec:cstp}). We have discussed the implementation of our games and the mechanism we set in place to attract users (Section~\ref{sec:game}). We detailed early results in the form usage statistics (Section~\ref{sec:stats}). 

While we are still in early stages of development and advertisement, the data we have already collected suggest that while the majority of word pairs can be reliably distinguished, some pairs may indeed prove to be challenging.
To give an anecdotal overview of what such pairs could be, we can highlight that in all the following word pairs the target has been correctly identified only 2 out of 6 times: Spanish ``cilantro'' and ``cebollino'', French ``chenapan'' and ``polisson'', English ``hyena'' and ``jackal''.
These encouraging results strongly suggest our experimental design has intrinsic worth: for instance, it has already reliably demonstrated how user-provided word pairs are generally less challenging than manually constructed ones.

On the other hand, further data collection is required before any strong conclusion can be reached. 
In future work, we intend to focus more keenly on the data we collect to identify hard-to-distinguish word pairs, as well as study the impact of the various strategies used to create word pairs.
We also plan to compare human behavior to that of large language models and distributional models, both semantically grounded and ungrounded.

\section*{Acknowledgment}
We thank Maria Copot, Herm\`es Martinez, Elena del Olmo, Tomara Gotkova and Aurore Coince for helping translating and advertising our game, Cyril Pestel for his help in deploying the game online, and Kar\"en Fort, Mathieu Lafourcade and three anonymous reviewers for suggestions for improvement. 
This work was supported  by a public grant overseen by the French National Research Agency (ANR) as part of the ``Investissements d'Avenir'' program: Idex \emph{Lorraine Universit\'e d'Excellence} (reference: ANR-15-IDEX-0004).






\end{document}